\documentclass{amia}
\usepackage{lipsum} 
\usepackage{amsmath}
\usepackage{algpseudocode}
\usepackage{algorithm}
\usepackage{graphicx}
\usepackage{adjustbox}
\usepackage{color}
\usepackage{enumitem}
\usepackage{makecell}
\usepackage{tabularx}
\usepackage{booktabs} 
\usepackage{array}    
\usepackage{caption}  
\usepackage{hyperref}

\setlist{topsep=0pt, leftmargin=*}

\begin{document}

\title{Integrating Social Determinants
of Health into Knowledge Graphs: Evaluating Prediction Bias and Fairness in Healthcare}

\author{
Tianqi Shang, MS$^{1,*}$, Weiqing He, MS$^{1,*}$, \\Tianlong Chen, PhD$^{2}$, Ying Ding$^{3}$, Huanmei Wu$^{4}$, Kaixiong Zhou, PhD$^{2}$, Li Shen, PhD$^{1,\dag}$}

\def\thefootnote{${*}$}\footnotetext{These authors contributed equally to this work. $^\dag$Correspondence to li.shen@pennmedicine.upenn.edu.}

\institutes{
    $^1$ Unversity of Pennsylvania, Philadelphia, PA, USA \\
    $^2$ University of North Carolina at Chapel Hill, Chapel Hill, NC, USA \\
    $^3$ University of Texas at Austin, Austin, TX, USA \\
    $^4$ Temple University, Philadelphia, PA, USA
}

\maketitle

\section*{Abstract}
\textit{Social determinants of health (SDoH) play a crucial role in patient health outcomes, yet their integration into biomedical knowledge graphs remains underexplored. This study addresses this gap by constructing an SDoH-enriched knowledge graph using the MIMIC-III dataset and PrimeKG. We introduce a novel fairness formulation for graph embeddings, focusing on invariance with respect to sensitive SDoH information. Via employing a heterogeneous-GCN model for drug-disease link prediction, we detect biases related to various SDoH factors. To mitigate these biases, we propose a post-processing method that strategically reweights edges connected to SDoHs, balancing their influence on graph representations. This approach represents one of the first comprehensive investigations into fairness issues within biomedical knowledge graphs incorporating SDoH. Our work not only highlights the importance of considering SDoH in medical informatics but also provides a concrete method for reducing SDoH-related biases in link prediction tasks, paving the way for more equitable healthcare recommendations. Our code is available at \url{https://github.com/hwq0726/SDoH-KG}.}
\section*{Introduction}



Knowledge graph is widely used to organize real-world entities and their relationships in a graph structure~\cite{fensel2020introduction, zou2020survey}. It is comprised of nodes and edges, where nodes can represent entities like objects, events, concepts, and so on, while edges define the relationships between them. 
Knowledge graphs have emerged as powerful tools in biomedical research~\cite{mohamed2021biological}\cite{carvalho2023knowledge}, enabling the representation of complex relationships between various biomedical entities such as drugs, diseases, and genes. For example, PrimeKG~\cite{chandak2023building}, a multimodal knowledge graph for precision medicine analyses; Monarch~\cite{putman2024monarch}, an analytic platform integrating phenotypes, genes, and diseases across species; DRKG~\cite{ioannidis2020few}, a comprehensive biological knowledge graph designed for Covid-19 drug repurposing. By structurally storing the vast amounts of biomedical knowledge triplets, these graphs have facilitated the advancements in drug discovery, disease prediction, personalized medicine, and so on. 

Recently, machine learning (ML) models have been developed to capture the complex structure of knowledge graphs, with a focus on tasks such as link prediction, node classification, and graph completion \cite{yang2014embedding, zhou2020multi, zhou2022auto, zhou2021temporal, hu2020open}. The existing effective knowledge graph approaches can be roughly divided into two groups: traditional embedding and deep neural networks.  First, the traditional knowledge graph embedding models like TransE~\cite{bordes2013translating}, TransR~\cite{lin2015learning}, and RotatE~\cite{sun2019rotate}, map the entities and relations within a knowledge graph into low-dimension vectors in a continuous embedding space, making it easier to perform downstream tasks. 
Second, Graph Neural Networks (GNNs)~\cite{scarselli2008graph} have been explored to process graph-structured data and encode entity features. GNNs work by iteratively aggregating information from a node's neighbors to learn node representations. Popular variants include Graph Convolutional Networks (GCNs)~\cite{kipf2016semi}, Graph Attention Networks (GATs)~\cite{velivckovic2017graph}, and Relational GCNs (R-GCNs)~\cite{schlichtkrull2018modeling}. Besides the modeling advancements, the quality of graph data itself play a crucial role for the accuracy and trustworthiness of many downstream healthcare tasks, e.g., discovering potential drug-disease associations or predicting treatment outcomes.  The edge connections among the biomedical entities will significantly influence the aggregation learning, node embeddings, and thus decision outcomes. Although GNNs widely deliver prediction efficacy in healthcare applications, they notoriously tend to capture structural bias and lead to  biased predictions. There exists a research gap in how to bridge GNNs and healthcare knowledge graph construction. 


Social Determinants of Health (SDoH), including the conditions in which people are born, grow, live, work, and age, as well as the broader forces and systems that shape daily life, are non-medical factors that impact health outcomes~\cite{healthypeople}. The impact of SDoH on health disparities has been a growing area of focus, with substantial research dedicated to understanding and addressing their influence~\cite{braveman2011social}\cite{donkin2018global}. While incorporating SDoH data into predictive models can enhance accuracy, fairness issues may arise when certain groups—based on socioeconomic status, education, race, or other factors—are underrepresented or misrepresented in the data~\cite{li2022improving}\cite{wang2019using}\cite{paulus2020predictably}. In biomedical applications, such biases can disproportionately affect marginalized populations, exacerbating health disparities. For instance, insurance companies might use algorithms informed by SDoH data to assess risk and determine coverage. Individuals from areas with high rates of unemployment might be considered at higher risk due to potential gaps in health coverage or inconsistent medical care, which could affect the premiums they pay or their eligibility for certain plans. The inclusion of SDoH factors in knowledge graphs, while essential for understanding the broader context of health outcomes, also introduces the risk that machine learning models trained on these graphs may propagate or even amplify existing inequities in healthcare delivery and treatment.  As machine learning continues to drive decision-making in healthcare, ensuring fairness in models trained on knowledge graphs is becoming increasingly critical. Addressing bias in these systems is essential not only for the accuracy of predictions but also for their ethical deployment in medical practice. However, while the fairness of ML models has been studied extensively in various fields~\cite{mehrabi2021survey}\cite{chouldechova2018frontiers}\cite{friedler2019comparative}, little research has been conducted on the fairness implications of SDoH in knowledge graphs. This brings us to the core focus of this work: investigating how SDoH data in biomedical knowledge graphs affects link prediction results, and proposing methodologies to detect and mitigate bias, ultimately advancing fairness in healthcare AI.

In this work, we collect patients' disease, drug, and SDoH information from the MIMIC-III dataset and phenotype information from PrimeKG to construct an SDoH knowledge graph. Based on this constructed graph,  we employ the idea of demographic parity~\cite{mehrabi2021survey, dwork2012fairness, kusner2017counterfactual} to develop a novel formulation of fairness within the context of link prediction, focusing on the invariance with respect to sensitive SDoHs. Specifically, we train a heterogeneous-GCN model to perform link prediction between drugs and diseases, and use this fairness notion to detect bias related to different types of SDoH information. For example, we explore whether the economics or education information can affect the medication recommendations between certain drug and disease. We also proposed a post-processing method to eliminate this bias. The key idea behind our de-biasing method is the strategic re-weighting of edges connected to SDoHs. This re-weighting aims to balance the influence of different sensitive SDoH factors, reducing the risk of unfair disparities in the graph's representation and subsequent predictions. To the best of our knowledge, this study represents one of the first comprehensive investigations into SDoH fairness issues within biomedical knowledge graphs.

\section*{Preliminaries}
In this section, we sequentially introduce the concepts of \textit{Heterogeneous Graphs}, \textit{Graph Convolutional Networks (GCNs)}, and \textit{link prediction}, providing foundational context for the subsequent discussions.
\subsection*{Heterogeneous Graph}
A heterogeneous knowledge graph is a type of graph that contains multiple types of nodes and/or multiple types of edges. This is in contrast to a homogeneous graph, where all nodes and edges are of the same type. In a heterogeneous graph, the different node types and edge types capture more complex relationships and interactions between various entities. We consider a heterogeneous graph $\mathcal{G}=(\mathcal{V}, \mathcal{E})$, which consists of a set of directed edge triples $e=(u, r, v) \in \mathcal{E}$, where $u, v \in \mathcal{V}$ are nodes and $r \in \mathcal{R}$ is a relation type. We further assume that each node is of a particular type, $\mathcal{T} \subseteq \mathcal{V}$, and that relations may have constraints regarding the types of nodes that they can connect. For example, in a heterogeneous biomedical knowledge graph, nodes include different entities like drugs, diseases, and genes, while edges represent different types of relationships, such as "treats," "causes," or "is associated with."
\subsection*{Graph Convolutional Networks} \label{GCN}
Graph Convolutional Networks (GCNs) is a type of neural network designed to operate directly on graph-structured data. It extends the concept of convolution from grid-like data (such as images) to arbitrary graphs, allowing node features to be aggregated from their neighbors, thus capturing the local structure of the graph. In a GCN, each node's representation is updated by aggregating feature information from its neighbors, weighted by the graph structure. The transformation at each layer can be expressed as:
\begin{equation}
    h_i^{(l+1)}=\sigma\left(b^{(l)}+\sum_{j \in \mathcal{N}(i)} \frac{1}{c_{j i}} h_j^{(l)} W^{(l)}\right)
\end{equation}
where $\mathcal{N}(i)$ is the set of neighbors of node $i, c_{j i}$ is the product of the square root of node degrees (i.e., $c_{j i}=\sqrt{|\mathcal{N}(j)|} \sqrt{|\mathcal{N}(i)|}$ ), $b^{(l)}$ is a bias term added to adjust the output, and $\sigma$ is an activation function (e.g., ReLU). This formulation enables the GCN to learn node representations by leveraging both node features and the graph's structure. Mathematically, GCN maps each node $v \in \mathcal{V}$ to a fixed size vector $\mathbf{z}_v$, and $\mathbf{z}_v$ can be used for downstream tasks like node classification, link prediction, and graph classification.
\subsection*{Link Prediction}
Link prediction is a task in graph analysis aimed at predicting the likelihood of potential or missing connections between nodes in a graph. The link prediction task on graph $\mathcal{G}$ can be defined as follows. Let $\mathcal{E}_{\text {train }} \subset \mathcal{E}$ denotes a set of observed training edges and let $\hat{\mathcal{E}}=\{\left( v_i, r, v_j\right): v_i, v_j \in\mathcal{V}, r \in R\} \backslash \mathcal{E}$ denotes the set of negative edges that are not present in the true graph $\mathcal{G}$. Given $\mathcal{E}_{\text {train }}$ and a scoring function $\phi$, an ideal model should assign higher scores to positive (true) edges than to any negative (false) edges. Specifically, if we focus at the link prediction of relationship $r^*$, using GCN as the embedding model and dot product as the scoring function, then we expect:
\begin{equation}
    \langle \mathbf{z}_v, \mathbf{z}_u \rangle > \langle \mathbf{z}_{v'}, \mathbf{z}_{u'} \rangle, \;\forall (u, r^*, v) \in \mathcal{E}, (u', r^*, v') \in \hat{\mathcal{E}}
\label{score}
\end{equation}
In the training phase, for each positive nodes pair (edge) $(u,v)$, in the target relationship $r^*$, we sample $K$ arbitrary nodes pairs $\{(u,v_i)\}_{i=1}^K$. According to equation~\ref{score}, we encourage the score between node $u$ and $v$ to be higher than the score between node $u$ and a sampled node $v_i$, and use \textit{margin loss} as the loss function to train a model for link prediction task,
\begin{equation}
    \mathcal{L}=\sum_{i=1,\ldots, K} \max \left(0, 1-\phi(u,r^*, v)+\phi(u,r^*, v_i)\right)
\end{equation}
The training process is end-to-end, where the embedding model, such as a GCN, is trained to generate node representations that minimize the margin loss.
\section*{Research Design and Methods}  
\begin{figure}[h]
    \centering
    \includegraphics[width=0.9\textwidth]
    {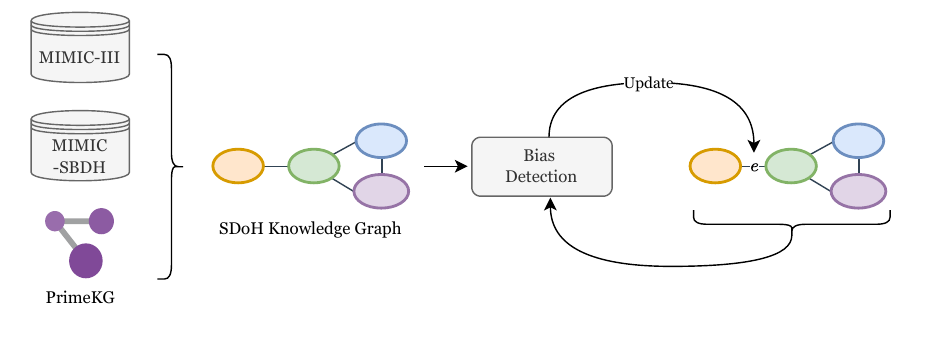}
    \vspace{-0.3cm}
    \caption{\small The workflow of this study. We first construct the SDoH knowledge graph by integrating MIMIC-III, MIMIC-SBDH, and PrimeKG. Then we define the fairness notion and detect the bias with respect to different sensitive SDoH. Finally, we propose a re-weighting strategy to mitigate bias by learning a weighting parameter on the edges between SDoH and drugs.}
    \label{fig:workflow}
\end{figure}

In this section, we begin by explaining the construction of the SDoH knowledge graph, detailing the types of nodes and the relationships captured in the graph. Next, we motivate and argue in favor of a particular form of "fairness" within the context of graph link prediction. Following this, we outline our de-biasing method which involves a re-weighting strategy on the edges. Figure \ref{fig:workflow} presents our workflow.

\subsection*{Dataset Construction}
MIMIC-SBDH~\cite{alsentzer2019publicly}\cite{ahsan2021mimic} is a data set containing 7,025 discharge summary notes randomly selected from the MIMIC III dataset~\cite{johnson2016mimic}. The notes are annotated for the patient’s status of the following Social and Behavioral Determinants of Health (SBDHs): Community, Education, Economics, Environment, Alcohol Use, Tobacco Use, and Drug Use, which we consider as Social Determinants of Health (SDoH) in this study.

For each discharge summary in MIMIC-SBDH, we identify its corresponding patient in the MIMIC III dataset according to its note ID. This allows us to obtain the patient's diagnosis and prescription data, capturing the disease and drug information of the patient. By linking these three entity types—disease, drug, and SDoH—we established relationships that provide a comprehensive view of each patient’s health profile. To further extend this data, we incorporated phenotype information from PrimeKG. PrimeKG provides a more holistic representation, as it directly connects phenotypes with both drugs and diseases. Since phenotypes are more likely to be influenced by SDoH factors than genetic information alone, incorporating phenotype data allowed us to enhance the dataset’s representation of the social and biological determinants of health~\cite{who2008closing}.

\begin{figure}[h]
    \centering
    \includegraphics[width=0.6\textwidth]
    {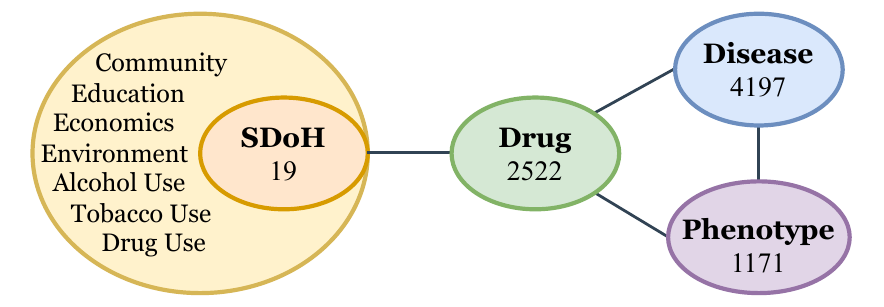}
    \vspace{-0.3cm}
    \caption{\small Architecture of our knowledge graph and the number of each type of node.}
    \label{fig:architecture}
\end{figure}

As outlined in Table~\ref{tab:sdoh}, each SDoH category contains several values describing the patient's status. Social supports such as family and friends are the context of Community, it is split into Community-Present and Community-Absent here, recognizing that a patient can have active social support (from family and friends) while simultaneously experiencing the loss of social support (due to events like death, separation, or divorce).  In MIMIC-SBDH, an SDoH is labeled \textit{True} when the discharge summary contains relevant information and \textit{None} if there is no such content. The \textit{False} label indicates either the opposite of \textit{True} (for example, homelessness under the environment category) or the absence of specific community support content. For alcohol, tobacco, and drug, the dataset captures consumer status using \textit{Present}, \textit{Past}, or \textit{Never}, while \textit{Unsure} is annotated when there are ambiguous contents in the discharge summary. We exclude all the 'None' and 'Unsure' values to ensure that only the useful SDoH information is retained and use the rest as an SDoH node in the knowledge graph. Thus, we can construct an SDoH knowledge graph with a total of 1,117,179 edges and 7,909 nodes, the structure and node distribution are shown in Figure~\ref{fig:architecture}.

\vspace{0.3cm}

\begin{table}[ht]
\caption{\small SDoH defined by MIMIC-SBDH and the values used in our study}
\vspace{-0.3cm}
\centering
\resizebox{0.8\textwidth}{!}{%
\begin{tabular}{lll}
\toprule
\multicolumn{1}{c}{\textbf{SDoH Category}} & \multicolumn{1}{c}{\textbf{Original Values}} & \multicolumn{1}{c}{\textbf{Values for this Study}} \\ \midrule
Community-Present& 0: False, 1: True                                 & 0: False, 1: True             \\
Community-Absent& 0: False, 1: True                                 & 0: False, 1: True             \\
Education   & 0: False, 1: True                                 & 0: False, 1: True             \\
Economics   & 0: None, 1: True, 2: False                        & 1: True, 2: False             \\
Environment & 0: None, 1: True, 2: False                        & 1: True, 2: False             \\
Alcohol Use & 0: None, 1: Present, 2: Past, 3: Never, 4: Unsure & 1: Present, 2: Past, 3: Never \\
Tobacco Use & 0: None, 1: Present, 2: Past, 3: Never, 4: Unsure & 1: Present, 2: Past, 3: Never \\
Drug Use    & 0: None, 1: Present, 2: Past, 3: Never, 4: Unsure & 1: Present, 2: Past, 3: Never \\ \bottomrule
\end{tabular}%
}

\label{tab:sdoh}
\end{table}



\subsection*{Fairness Notion and Bias Detection}
In the section, we discuss the definition of our fairness notion and the way we detect the bias. We consider a simple, intuitive formulation of fairness within the context of link prediction. Using economic level as an example of a sensitive SDoH and drug re-purposing as an example relation prediction task, our approach is guided by the following question: \textit{What potential new therapeutic uses might be identified for an existing drug if we disregard the economic information associated with it?} In other words, the discovery would be the same regardless of the economics information. Formally, given a drug $u$ and a disease $v$, for the edge between $u$ and $v$, the score (probability) should be independent with sensitive SDoH information $\mathcal{T}^*$,
\begin{equation}
    \phi(\mathbf{z}_u, r, \mathbf{z}_v) \perp \mathbf{1}_{(u,w)}, \;\forall r\in \mathcal{R}', w\in \mathcal{T}^*
\end{equation}
where $\mathbf{1}_{(u,w)}$ is the indicator function, equals to 1 if there is an edge between $u$ and $w$, and 0 otherwise, and $\mathcal{R}'$ denotes the set of all possible relations between drugs and disease. For simplicity, we consider only one possible relation from drug to disease (i.e., $|\mathcal{R'}| = 1$), and we can abbreviate $\phi(\mathbf{z}_u, r, \mathbf{z}_v)$ as $\phi(\mathbf{z}_u, \mathbf{z}_v)$. Additionally, we assume that the set $\mathcal{T}^*$ contains only two nodes\footnote{This can be extended for more than two nodes.}, denoted as $w_0$ and $w_1$.  With this in mind, we define the bias with sensitive SDoH factor $\mathcal{T}^*$ as:
\begin{equation}
    D_{\mathcal{T}^*} = \Big| \phi[(\mathbf{z}_u, \mathbf{z}_v) \mid \mathbf{1}_{(u,w_0)} = 1] - \phi[(\mathbf{z}_u, \mathbf{z}_v) \mid \mathbf{1}_{(u,w_1)} = 1] \Big|
\end{equation}
$ D_{\mathcal{T}^*}$ evaluates the disparity in the score of an edge between drug $u$ and disease $v$ when $u$ associates with different sensitive SDoH information. It is worth mentioning that our fairness notion is similar to traditional demographic parity~\cite{mehrabi2021survey, dwork2012fairness, kusner2017counterfactual}, as both are independent of the true label (i.e., they do not consider whether the edge actually exists). However, in traditional fairness problems in knowledge graphs, sensitive information is typically encoded as node attributes with varying values. In contrast, in our approach, we treat sensitive information as separate nodes and examine bias based on different connection patterns.

Next we apply this notion to detect the bias with respect to different potential sensitive SDoH factors. We begin with the definition of \textit{$\mathcal{T}_i$-free drug nodes}. As mentioned before, we have eight categories of SDoHs and we consider Community Present/Absent, Education, Economics, and Environment as the potential sensitive SDoHs, denoted as $\{\mathcal{T}_i\}_{i=1}^5$. We then define \textit{$\mathcal{T}_i$-free drug nodes} as the drug nodes which are not connected with any nodes in $\mathcal{T}_i$. For example, drug \texttt{481(*IND* Pexelizumab/Placebo)} is not connected with either 'Environment: True' or 'Environment: False', then it is considered an Environment-free drug node. In other words, the $\mathcal{T}_i$-free drug nodes contain no information of $\mathcal{T}_i$. 

\begin{figure}[h]
    \centering
    \includegraphics[width=\textwidth]
    {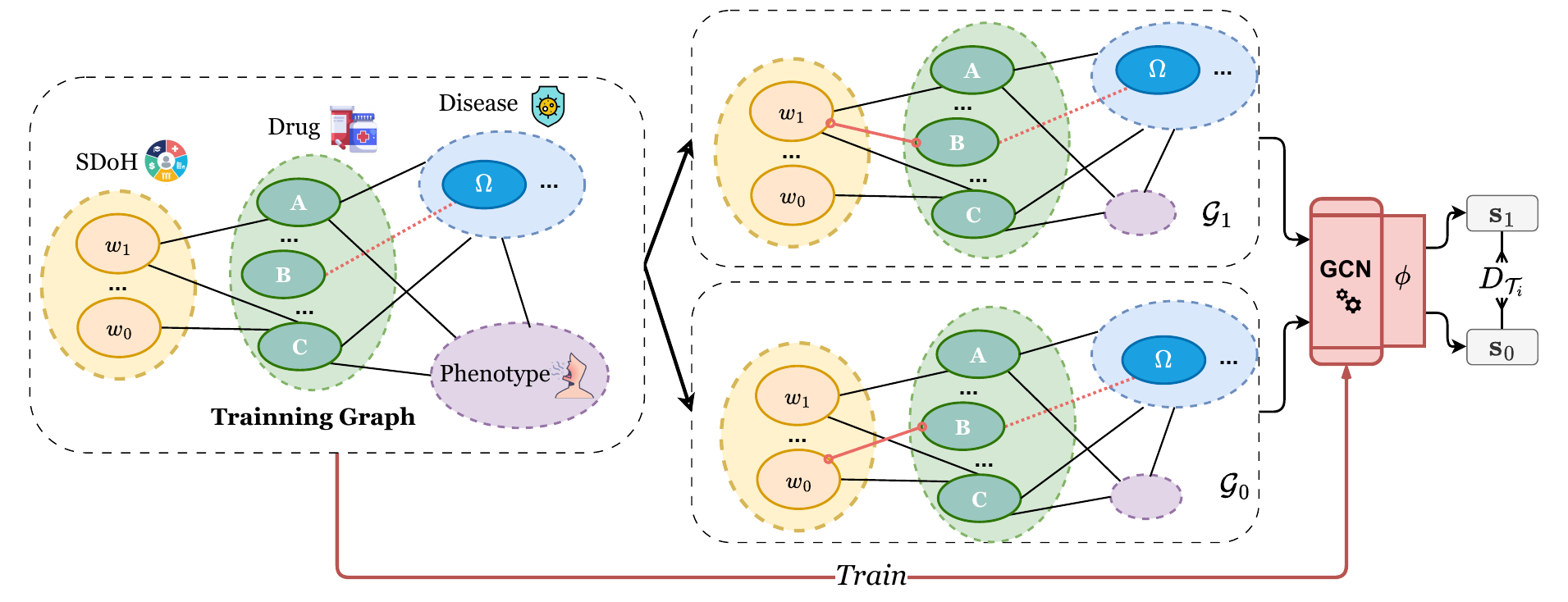}
    \vspace{-0.3cm}
    \caption{\small The workflow for bias detection. This figure illustrates the process of detecting bias with respect to sensitive SDoH $\mathcal{T}_i = \{w_0, w_1\}$. In the figure, drug nodes \textbf{B} is a \textit{$\mathcal{T}_i$-free drug node}, and we mask the edge with disease $\Omega$ to build the training graph. We connect \textbf{B} with SDoH $w_0$ and $w_1$ respectively to construct two testing graphs: $\mathcal{G}_0$ and $\mathcal{G}_1$. A GCN trained on the training graph is then used to perform inference on both testing graphs, and the resulting scores are compared to detect bias.}
    \label{fig:detectbias}
\end{figure}

To detect the bias with respect to sensitive SDoH $\mathcal{T}_i = \{w_0, w_1\}$, we first collect $\mathcal{T}_i$-free drug nodes and for each of these nodes choose one drug-disease edge to build the test edge set $\mathcal{C}$. Then we mask all the edges in $\mathcal{C}$ to generated a training graph. Using this training graph to create two testing graphs: $\mathcal{G}_0$ by connecting $\mathcal{T}_i$-free drug nodes to $w_0$ and masking the edges in the test edge set, and $\mathcal{G}_1$ by connecting $\mathcal{T}_i$-free drug nodes to $w_1$ and similarly masking the test edges. We then train a GCN model on the training graph, use the trained model to perform inference on $\mathcal{G}_0$ and $\mathcal{G}_1$ respectively. Employing dot product as the scoring function, we compute the scores $\mathbf{s}_0$ and $\mathbf{s}_1$ for the test edges on each respective graph. Finally, we get the bias $D_{\mathcal{T}_i}$ by calculating the mean absolute difference between the scores $\mathbf{s}_0$ and $ \mathbf{s}_1$. Figure~\ref{fig:detectbias} illustrates this detection process.


\subsection*{De-bias Method}
\begin{figure}[t]
    \centering
    \includegraphics[width=\textwidth]
    {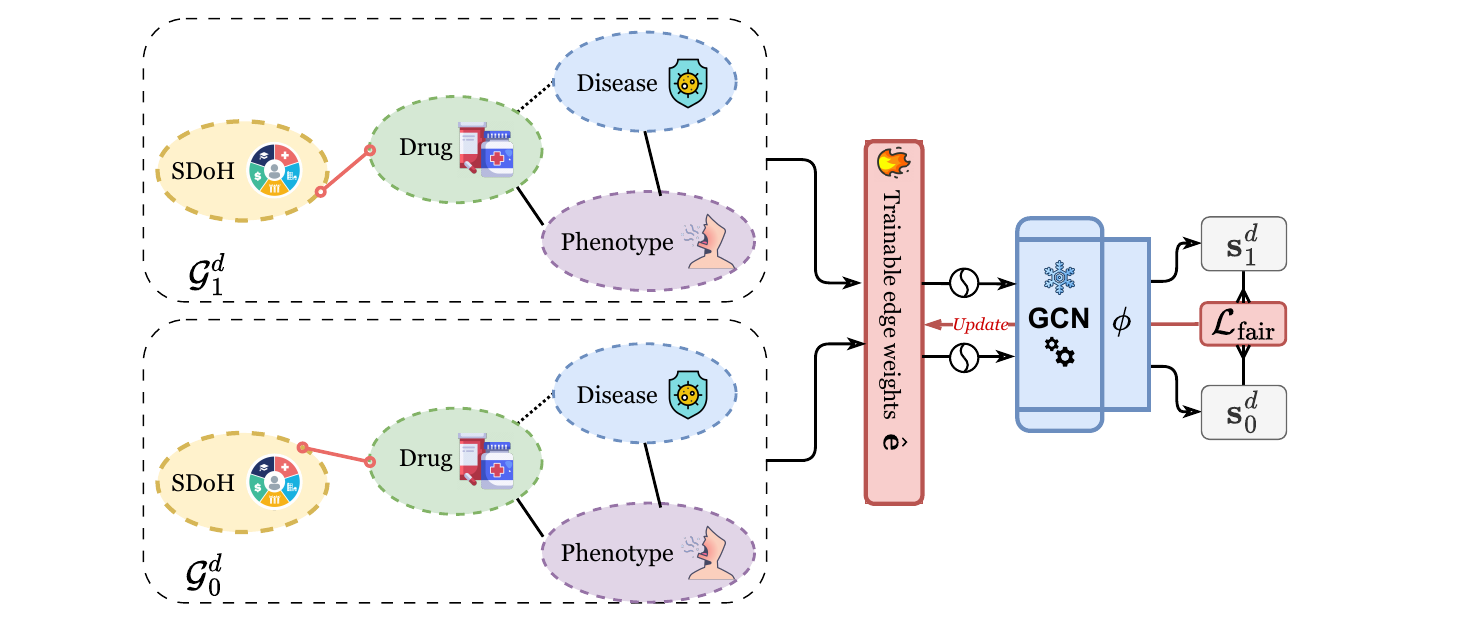}
    \vspace{-0.3cm}
    \caption{\small The training process of the de-biasing method. We first follow the steps in Bias Detection but collect a different test edge set to generate two training graphs for de-biasing: $\mathcal{G}_0^d, \mathcal{G}_1^d$. Then, with the weighted-GCN, we perform inference on both of the graphs and compute the loss. During the update phase, all the parameters in pre-trained GCN are frozen and only the weighting parameter $\hat{\mathbf{e}}$ is updated to minimize the loss.}
    \label{fig:debias}
\end{figure}

In this work, our fairness-aware method aims at removing bias from the graph structure itself by re-weighting the edges between drug and SDoHs. As mentioned in the Preliminaries, the weight of each edge in the graph is all considered equally, in other words, the graph's adjacency matrix only contains 0 and 1. By assigning equal weight to all edges during aggregation, the model can unintentionally amplify biases in the data. In a knowledge graph containing multiple types of SDoHs, this equal treatment assumes that all edges—representing relationships like economic level or alcohol use—contribute equally to the predictions. However, sensitive SDoH like economics may have disproportionate influence on medical outcomes, leading to biased predictions. For instance, individuals from lower economic levels may face different healthcare challenges, which could skew the predictions if those edges are not properly weighted. By treating every edge equally, the model risks over-representing or under-representing certain SDoH factors, leading to unfair and biased link predictions between drugs and diseases. To address this inherent bias in traditional GCNs, we proposed a re-weighting strategy on the edges between SDoHs and drugs. We begin with adding weighting parameter to each layer of GCN, the weighted transformation at each layer is expressed as:
\begin{equation}
    h_i^{(l+1)}=\sigma\left(b^{(l)}+\sum_{j \in \mathcal{N}(i)} \frac{e_{ji}}{c_{j i}} h_j^{(l)} W^{(l)}\right)
\label{weighted-GCN}
\end{equation}
where $e_{ji}$ is the scalar weight on the edge from node $j$ to node $i$ and other parameters remain consistent with those defined in the Preliminaries. The involvement of the weighting parameter enables the model to assign different importance to each edge and allows for a more nuanced representation of relationships by controlling the contribution of sensitive SDoHs, such as economic level, in the prediction process. During the training process, we aim to learn the weighting parameter that minimizes bias with respect to sensitive SDoH factors. To detect and remove the bias with respect to sensitive SDoH $\mathcal{T}_i = \{w_0, w_1\}$, we follow the steps in Bias Detection but collect a different test edge set $\mathcal{C}'$ to generate two training graphs for de-biasing: $\mathcal{G}_0^d, \mathcal{G}_1^d$. For the pre-trained GCN model, we freeze all the learnable parameters except for the initialized weighting parameter $\hat{\mathbf{e}}$ (i.e., in equation~\ref{weighted-GCN}, we freeze $b^{(l)}$ and $W^{(l)}$). Using the weighted-GCN, we perform inference on $\mathcal{G}_0^d$ and $\mathcal{G}_1^d$. With the dot product as the scoring function, we compute the scores of test edge set $\mathcal{C}'$ on both graphs, denoted as $\mathbf{s}^d_0, \mathbf{s}^d_0$. The loss function is then defined as the mean squared difference between these two scores:
\begin{equation}
    \mathcal{L}_\text{fair} = \frac{1}{n} \sum_{i=1}^n\left(s_{0, i}^d-s_{1, i}^d\right)^2
\end{equation}
where $n$ is the length of the vectors (the number of elements in $\mathbf{s}^d_0$ and $\mathbf{s}^d_1$), and $s^d_{0,i}, s^d_{1,i}$ are the $i$-th elements of the vectors $\mathbf{s}^d_0$ and $\mathbf{s}^d_1$ respectively. In the training process, only the weighting parameter $\hat{\mathbf{e}}$ is learnable, and with back-propagation $\hat{\mathbf{e}}$ can be updated to minimize the loss. Figure~\ref{fig:debias} presents the training process of the de-biasing method.
\section*{Results}

As mentioned in the previous section, we limit our fairness analysis to the following categories: Education, Economics, Environment, and Community Present/Absent. Alcohol, tobacco, and drug use are not considered as sensitive factors in this context since fairness evaluations on these categories would not provide meaningful insights. Sensitive attributes in fairness evaluations are typically those that are associated with systemic inequalities or discrimination, such as race, gender, socioeconomic status, or education level. These attributes are protected under various ethical and legal frameworks (e.g., anti-discrimination laws like the U.S. Civil Rights Act or the European Union's General Data Protection Regulation), which aim to prevent models from disproportionately impacting specific demographic groups. In contrast, alcohol, tobacco, and drug use are considered behavioral factors that do not inherently align with these protected categories. While these behaviors may influence health outcomes, they are generally viewed as personal lifestyle choices rather than attributes that represent underlying social inequalities. Furthermore, individuals may engage in or abstain from these behaviors for a variety of reasons—such as personal preference, medical advice, or cultural norms—which complicates their classification as sensitive attributes. For the selected five SDoH categories, we applied the bias detection and de-biasing method as outlined in the Research Design and Methods section to measure and mitigate bias.

As shown in Table \ref{tab:result}, our de-biasing method significantly reduced the bias across all five SDoH categories. The bias values in the table represent the disparity in prediction scores associated with different SDoH factors, as defined by our fairness notion. The "Initial Bias" column indicates the extent of bias before applying our de-biasing method, while the "Current Bias" shows the reduced bias after applying our re-weighting strategy on the edges between SDoHs and drugs. The reduction in all five SDoH categories proves that our method is both effective and comprehensive. In addition to measuring bias reduction, we evaluated the impact of our de-biasing method on the predictive performance of the model, as indicated by the Mean Reciprocal Rank (MRR) which evaluates the mean reciprocal rank of the positive edge score among the arbitrary edge scores (in our experiment, we samlpe 20 arbitrary nodes pairs for each positive nodes pair.). The original MRR was 0.36064, and after de-biasing, the MRR values for each category were marginally different, ranging between 0.36066 and 0.36071. These slight variations indicate that the de-biasing process did not significantly alter the model's predictive accuracy. The minimal change in MRR demonstrates that our de-bias method strikes a crucial balance between fairness and accuracy. Achieving fairness often comes at the cost of reduced model performance, but in our case, the model retained its high accuracy while also becoming significantly fairer across multiple sensitive attributes. This indicates that our de-biasing method is both efficient and practical for real-world biomedical applications, where fairness and accuracy are equally important.
\vspace{0.3cm}
\begin{table}[ht]
\caption{The change of \textit{bias} and \textit{MRR} after de-biasing (lower \textit{bias} is better, higher \textit{MRR} is better)}
\vspace{-0.3cm}
\centering
\resizebox{0.9\textwidth}{!}{%
\begin{tabular}{lrrrrc}
\toprule
\multicolumn{1}{c}{\textbf{SDoH Category}} &
  \multicolumn{1}{c}{\textbf{Initial Bias}} &
  \multicolumn{1}{c}{\textbf{Current Bias}} &
  \multicolumn{1}{c}{\textbf{Bias Reduction}} &
  \multicolumn{1}{c}{\textbf{Current MRR}} &
  \multicolumn{1}{c}{\textbf{MRR Difference}} \\ \midrule
Education         & 67.95  & 8.83  & 59.12  & 0.36071 & -6.58e-05 \\
Economics         & 92.40  & 6.91  & 85.49  & 0.36066 & -1.93e-05 \\
Environment       & 158.30 & 20.36 & 137.94 & 0.36068 & -4.31e-05 \\
Community Present & 5.30   & 1.37  & 3.93   & 0.36071 & -7.13e-05 \\
Community Absent  & 33.16  & 3.15 & 30.01   & 0.36069 & -5.55e-05 \\ \bottomrule
\end{tabular}
}

\label{tab:result}
\end{table}

\section*{Discussion and Implications}

\paragraph{\textit{Contributions}} 
This study makes several significant contributions to the field of biomedical informatics and fairness in machine learning. Firstly, we have successfully constructed a comprehensive knowledge graph that integrates Social Determinants of Health (SDoH) with traditional biomedical data, providing a more holistic representation of patient health factors. Secondly, we have introduced a novel fairness notion specifically tailored for graph embeddings, along with an innovative method to detect biases related to sensitive SDoH within the link prediction task. This advancement allows for a more nuanced understanding of how social factors may influence medical predictions and recommendations. Lastly, we have developed and presented a post-processing technique that effectively reduces SDoH-related biases while maintaining model accuracy, ensuring fairer and more equitable predictions in real-world biomedical applications, where fairness and accuracy are equally important.

\paragraph{\textit{Limitations and Future Work}} 
Despite the significant strides made in this study, several limitations warrant consideration and point towards directions for future research. The primary limitation lies in the dataset used, which, while comprehensive, may not fully represent the diversity of global populations and healthcare systems. Future work should focus on validating our approach across more diverse datasets to ensure generalizability. Additionally, while our bias detection and mitigation methods show promise, they currently focus on a limited set of SDoH factors. Expanding this to include a broader range of social determinants could provide even more comprehensive insights into healthcare disparities. Another avenue for future research involves the exploration of more sophisticated graph neural network architectures, where the re-weighting strategy may need to be refined or adapted to suit the increased complexity of the model. Lastly, investigating the real-world impact of our fairness-enhanced predictions on clinical decision-making processes could provide valuable insights into the practical applications of this work.

\paragraph{\textit{Conclusion}}  
In conclusion, this study represents a significant step forward in addressing fairness issues within biomedical knowledge graphs, particularly concerning the integration and impact of Social Determinants of Health. By constructing an SDoH-enriched knowledge graph, defining novel fairness metrics for graph link prediction task, and developing methods to detect and mitigate bias, we have laid the groundwork for more equitable medical predictions and recommendations. As healthcare system continues to move towards more personalized and data-driven approaches, the methods and insights presented in this study will be useful in ensuring that these advancements benefit all individuals equally, regardless of their social circumstances. This work opens up new possibilities for fair and comprehensive medication analytics, paving the way for more equitable and effective healthcare systems.

\subparagraph{Acknowledgments} This work was supported in part by the NIH grants U01 AG066833, U01 AG068057, R01 AG071470 and P30 AG073105. 

\makeatletter
\renewcommand{\@biblabel}[1]{\hfill #1.}
\makeatother

\small
\bibliographystyle{vancouver}
\bibliography{amia}  

\begin{thebibliography}{10}

\bibitem{fensel2020introduction}
Fensel D, {\c{S}}im{\c{s}}ek U, Angele K, Huaman E, K{\"a}rle E, Panasiuk O, et~al.
\newblock Introduction: what is a knowledge graph?
\newblock Knowledge graphs: Methodology, tools and selected use cases. 2020:1-10.

\bibitem{zou2020survey}
Zou X.
\newblock A survey on application of knowledge graph.
\newblock In: Journal of Physics: Conference Series. vol. 1487. IOP Publishing; 2020. p. 012016.

\bibitem{mohamed2021biological}
Mohamed SK, Nounu A, Nov{\'a}{\v{c}}ek V.
\newblock Biological applications of knowledge graph embedding models.
\newblock Briefings in bioinformatics. 2021;22(2):1679-93.

\bibitem{carvalho2023knowledge}
Carvalho RM, Oliveira D, Pesquita C.
\newblock Knowledge Graph Embeddings for ICU readmission prediction.
\newblock BMC Medical Informatics and Decision Making. 2023;23(1):12.

\bibitem{chandak2023building}
Chandak P, Huang K, Zitnik M.
\newblock Building a knowledge graph to enable precision medicine.
\newblock Scientific Data. 2023;10(1):67.

\bibitem{putman2024monarch}
Putman TE, Schaper K, Matentzoglu N, Rubinetti VP, Alquaddoomi FS, Cox C, et~al.
\newblock The Monarch Initiative in 2024: an analytic platform integrating phenotypes, genes and diseases across species.
\newblock Nucleic acids research. 2024;52(D1):D938-49.

\bibitem{ioannidis2020few}
Ioannidis VN, Zheng D, Karypis G.
\newblock Few-shot link prediction via graph neural networks for covid-19 drug-repurposing.
\newblock arXiv preprint arXiv:200710261. 2020.

\bibitem{yang2014embedding}
Yang B, Yih Wt, He X, Gao J, Deng L.
\newblock Embedding entities and relations for learning and inference in knowledge bases.
\newblock arXiv preprint arXiv:14126575. 2014.

\bibitem{zhou2020multi}
Zhou K, Song Q, Huang X, Zha D, Zou N, Hu X, et~al.
\newblock Multi-channel graph neural networks.
\newblock International Joint Conferences on Artificial Intelligence; 2020. .

\bibitem{zhou2022auto}
Zhou K, Huang X, Song Q, Chen R, Hu X.
\newblock Auto-gnn: Neural architecture search of graph neural networks.
\newblock Frontiers in big Data. 2022;5:1029307.

\bibitem{zhou2021temporal}
Zhou H, Tan Q, Huang X, Zhou K, Wang X.
\newblock Temporal augmented graph neural networks for session-based recommendations.
\newblock In: Proceedings of the 44th International ACM SIGIR conference on research and development in information retrieval; 2021. p. 1798-802.

\bibitem{hu2020open}
Hu W, Fey M, Zitnik M, Dong Y, Ren H, Liu B, et~al.
\newblock Open graph benchmark: Datasets for machine learning on graphs.
\newblock Advances in neural information processing systems. 2020;33:22118-33.

\bibitem{bordes2013translating}
Bordes A, Usunier N, Garcia-Duran A, Weston J, Yakhnenko O.
\newblock Translating embeddings for modeling multi-relational data.
\newblock Advances in neural information processing systems. 2013;26.

\bibitem{lin2015learning}
Lin Y, Liu Z, Sun M, Liu Y, Zhu X.
\newblock Learning entity and relation embeddings for knowledge graph completion.
\newblock In: Proceedings of the AAAI conference on artificial intelligence. vol.~29; 2015. .

\bibitem{sun2019rotate}
Sun Z, Deng ZH, Nie JY, Tang J.
\newblock Rotate: Knowledge graph embedding by relational rotation in complex space.
\newblock arXiv preprint arXiv:190210197. 2019.

\bibitem{scarselli2008graph}
Scarselli F, Gori M, Tsoi AC, Hagenbuchner M, Monfardini G.
\newblock The graph neural network model.
\newblock IEEE transactions on neural networks. 2008;20(1):61-80.

\bibitem{kipf2016semi}
Kipf TN, Welling M.
\newblock Semi-supervised classification with graph convolutional networks.
\newblock arXiv preprint arXiv:160902907. 2016.

\bibitem{velivckovic2017graph}
Veli{\v{c}}kovi{\'c} P, Cucurull G, Casanova A, Romero A, Lio P, Bengio Y.
\newblock Graph attention networks.
\newblock arXiv preprint arXiv:171010903. 2017.

\bibitem{schlichtkrull2018modeling}
Schlichtkrull M, Kipf TN, Bloem P, Van Den~Berg R, Titov I, Welling M.
\newblock Modeling relational data with graph convolutional networks.
\newblock In: The semantic web: 15th international conference, ESWC 2018, Heraklion, Crete, Greece, June 3--7, 2018, proceedings 15. Springer; 2018. p. 593-607.

\bibitem{healthypeople}
2030 HP. Social Determinants of Health; 2024.
\newblock Available from: \url{https://health.gov/healthypeople/objectives-and-data/social-determinants-health}.

\bibitem{braveman2011social}
Braveman P, Egerter S, Williams DR.
\newblock The social determinants of health: coming of age.
\newblock Annual review of public health. 2011;32(1):381-98.

\bibitem{donkin2018global}
Donkin A, Goldblatt P, Allen J, Nathanson V, Marmot M.
\newblock Global action on the social determinants of health.
\newblock BMJ global health. 2018;3(Suppl 1):e000603.

\bibitem{li2022improving}
Li Y, Wang H, Luo Y.
\newblock Improving fairness in the prediction of heart failure length of stay and mortality by integrating social determinants of health.
\newblock Circulation: Heart Failure. 2022;15(11):e009473.

\bibitem{wang2019using}
Wang H, Li Y, Ning H, Wilkins J, Lloyd-Jones D, Luo Y.
\newblock Using machine learning to integrate socio-behavioral factors in predicting cardiovascular-related mortality risk.
\newblock In: MEDINFO 2019: Health and Wellbeing e-Networks for All. IOS Press; 2019. p. 433-7.

\bibitem{paulus2020predictably}
Paulus JK, Kent DM.
\newblock Predictably unequal: understanding and addressing concerns that algorithmic clinical prediction may increase health disparities.
\newblock NPJ digital medicine. 2020;3(1):99.

\bibitem{mehrabi2021survey}
Mehrabi N, Morstatter F, Saxena N, Lerman K, Galstyan A.
\newblock A survey on bias and fairness in machine learning.
\newblock ACM computing surveys (CSUR). 2021;54(6):1-35.

\bibitem{chouldechova2018frontiers}
Chouldechova A, Roth A.
\newblock The frontiers of fairness in machine learning.
\newblock arXiv preprint arXiv:181008810. 2018.

\bibitem{friedler2019comparative}
Friedler SA, Scheidegger C, Venkatasubramanian S, Choudhary S, Hamilton EP, Roth D.
\newblock A comparative study of fairness-enhancing interventions in machine learning.
\newblock In: Proceedings of the conference on fairness, accountability, and transparency; 2019. p. 329-38.

\bibitem{dwork2012fairness}
Dwork C, Hardt M, Pitassi T, Reingold O, Zemel R.
\newblock Fairness through awareness.
\newblock In: Proceedings of the 3rd innovations in theoretical computer science conference; 2012. p. 214-26.

\bibitem{kusner2017counterfactual}
Kusner MJ, Loftus J, Russell C, Silva R.
\newblock Counterfactual fairness.
\newblock Advances in neural information processing systems. 2017;30.

\bibitem{alsentzer2019publicly}
Alsentzer E, Murphy JR, Boag W, Weng WH, Jin D, Naumann T, et~al.
\newblock Publicly available clinical BERT embeddings.
\newblock arXiv preprint arXiv:190403323. 2019.

\bibitem{ahsan2021mimic}
Ahsan H, Ohnuki E, Mitra A, You H.
\newblock MIMIC-SBDH: a dataset for social and behavioral determinants of health.
\newblock In: Machine Learning for Healthcare Conference. PMLR; 2021. p. 391-413.

\bibitem{johnson2016mimic}
Johnson AE, Pollard TJ, Shen L, Lehman LwH, Feng M, Ghassemi M, et~al.
\newblock MIMIC-III, a freely accessible critical care database.
\newblock Scientific data. 2016;3(1):1-9.

\bibitem{who2008closing}
on~Social Determinants~of Health WC, Organization WH.
\newblock Closing the gap in a generation: health equity through action on the social determinants of health: Commission on Social Determinants of Health final report.
\newblock World Health Organization; 2008.

\end{thebibliography}
\end{document}